\documentclass[
twocolumn,
]{ceurart}

\usepackage{listings}
\usepackage{graphicx}
\begin{document}

\copyrightyear{2024}
\copyrightclause{Copyright for this paper by its authors.
  Use permitted under Creative Commons License Attribution 4.0
  International (CC BY 4.0).}

\conference{ }

\title{Beyond Labels: A Self-Supervised Framework with Masked Autoencoders and Random Cropping for Breast Cancer Subtype Classification}

\author[2,3,4]{Annalisa Chiocchetti}[%
email=annalisa.chiocchetti@med.uniupo.it,
]

\address[1]{DISIT, Computer Science Institute,
University of Piemonte Orientale, 15121 Alessandria, Italy}

\author[1,4]{Marco Dossena}[%
email=marco.dossena@uniupo.it,
]
\cormark[1]
\fnmark[2]
\address[2]{Center for Translational Research on Autoimmune and Allergic Diseases (CAAD), Department of Health Sciences, University of Piemonte Orientale, 28100 Novara, Italy}

\address[3]{Interdisciplinary Research Center of Autoimmune Diseases (IRCAD), Department of Health Sciences, University of Piemonte Orientale, Novara, 28100, Italy}

\address[4]{Interdepartmental Research Center on Artificial Intelligence (AI@UPO), University of Piemonte Orientale, 15121 Alessandria, Italy}

\author[1,4]{Christopher Irwin}[%
email=christopher.irwin@uniupo.it,
]
\fnmark[2]

\author[1,4]{Luigi Portinale}[%
email=luigi.portinale@uniupo.it,
]

\cortext[1]{Corresponding author.}
\fntext[1]{Authors are listed in alphabetical order.}
\fntext[2]{Phd student enrolled in the National PhD in Artificial Intelligence for
Health and Life Sciences, XXXVIII cycle, Università Campus Bio-Medico, Roma.}

\begin{abstract}
This work contributes to breast cancer sub-type classification using histopathological images. We utilize masked autoencoders (MAEs) to learn a self-supervised embedding tailored for computer vision tasks in this domain. This embedding captures informative representations of histopathological data, facilitating feature learning without extensive labeled datasets. During pre-training, we investigate employing a random crop technique to generate a large dataset from WSIs automatically. Additionally, we assess the performance of linear probes for multi-class classification tasks of cancer sub-types using the representations learnt by the MAE. Our approach aims to achieve strong performance on downstream tasks by leveraging the complementary strengths of ViTs and autoencoders. We evaluate our model's performance on the BRACS dataset and compare it with existing benchmarks.
\end{abstract}


\maketitle

\section{Introduction}

Histopathological image analysis plays a critical role in disease diagnosis, particularly in cancer. Whole-slide images (WSIs) offer high-resolution views of entire tissue sections, enabling comprehensive evaluation by pathologists. However, manual analysis of WSIs is time-consuming and prone to inter-observer variability. Deep learning models have emerged as powerful tools to automate histopathological image analysis, offering the potential for faster, more consistent, and potentially more accurate diagnoses.
Given the high-resolution nature of WSIs, it is interesting to adopt a localized analysis approach. This involved extracting image patches and applying tasks like classification at the level of bag of tissue regions, employing a Multiple Instance Learning (MIL) framework \cite{mil}. Furthermore, the dimensions of the extracted patches facilitate the application of deep learning models for computer vision tasks, including Convolutional Neural Networks (CNN) and Vision Transformer (ViT) architectures, for various objectives such as tissue classification and cell segmentation.

CNNs have become the most used approach in this domain due to their ability to capture spatial relationships within images~\cite{cnn1,cnn3}. Architectures such as VGG, ResNet, and Inception excel at learning hierarchical features directly from raw image data, making them ideal for tasks like tissue classification, tumor segmentation, and cell detection.
However, CNNs struggle to capture long-range dependencies within complex tissues. 

This is where Graph Neural Networks (GNNs) offer a promising alternative. GNNs represent tissue structures as graphs \cite{gnn1, gnn3}, where nodes represent cells and edges depict their relationships. This allows GNNs to effectively model complex interactions between cells, making them particularly useful for analyzing cell-to-cell communication or studying the spatial distribution of different cell types.

Alternatively, for tasks aiming to classify multiple bag of regions under a single label, the graph representation can be constructed at the patch level. This approach treats each region as a node within the graph \cite{hact}.
ViTs represent another interesting approach. Unlike CNNs, ViTs process image patches directly, leveraging transformer techniques to learn global dependencies across the entire image~\cite{vit1, vit2}. 
This approach proves advantageous for tasks requiring analysis of intricate tissue patterns or overcoming limitations imposed by pre-defined filter sizes in CNNs.

Existing methods try to achieve an optimal latent representation that facilitates accurate classification of tissue regions or whole slide images.
We propose a reconstruction framework called \textit{Histopathological Masked AutoEncoder} (HMAE) followed by a simple classifier. The construction of the latent space of this model is achieved through the self-supervised objective of reconstructing the original image. By masking a significant portion of image patches during input, the encoder is forced to identify increasingly intricate patterns in the remaining data to reconstruct the complete image. Finally, a simple MLP with an output linear layer is applied after the training of the transformer, in order to perform classification, so that the model is never exposed to the image class labels during the learning process.

The datasets we have considered in the present work are the following.
\newline
\newline
\textbf{BRACS}. The BReAst Carcinoma Subtyping (BRACS) dataset~\cite{bracs} is a collection of digital images used to study breast lesions. It includes 547 WSIs, which are high-resolution scans of entire tissue samples. Additionally, 4539 smaller, more specific areas called regions of interest (ROIs) are extracted from these whole-slide images. Each WSI and its corresponding ROIs are carefully examined and labeled by three pathologists. These categories encompass three main lesion types: benign (healthy tissue), malignant (cancerous tissue), and atypical. Further details are provided by seven subcategories within these main types, allowing for a more precise understanding of the specific lesion.
\newline
\newline
\textbf{BACH}. The BreAst Cancer Histology (BACH) dataset~\cite{bach} is a significant resource for researchers developing computer algorithms to automatically diagnose breast cancer. It consists of a collection of digitized images from breast biopsies (WSIs). The ROIs extracted from each WSI are labeled according to four classifications: normal tissue, benign tumors, in situ carcinoma (precancerous cells), and invasive carcinoma (cancerous cells). There are 100 ROIs for each class.

\section{Methodologies}
This section investigates the synergy between ViTs and masking techniques for image reconstruction using a Masked Autoencoding framework (MAE). We will begin by investigating the theoretical foundations of ViT and masked image encoding, exploring their architectural details and functionalities. Next, we will describe the process of acquiring and pre-processing a dataset suitable for MAE training. This dataset will be extracted from whole slide images (WSI) obtained from BRACS patients. Finally, we will elucidate the methodology employed to leverage the feature representations (embeddings) learnt the ViT model for the task of cancer classification. 

\subsection{Vision Transformers}

Vision Transformers (ViTs) \cite{vit} represent a paradigm shift in computer vision, achieving state-of-the-art results on image classification tasks while departing from the traditional dominance of Convolutional Neural Networks (CNNs). Unlike CNNs that rely on hand-crafted filters for feature extraction, ViTs leverage the Transformer architecture, originally successful in Natural Language Processing (NLP).
The key idea is to split an image into fixed-size patches. These patches are then embedded into a vector representation and fed into a Transformer encoder. The Transformer utilizes self-attention \cite{attention} mechanisms to learn long-range dependencies between different parts of the image, allowing the model to capture global context crucial for classification.

\subsection{Masked Encoder}

As first, we divide the image into equally sized non-overlapping patches. Then, instead of using all the patches, we pick a certain number and hide the rest (masking) \cite{mae}. These patches are chosen randomly and without replacement. We also pick patches randomly across the whole image to avoid favoring the center (center bias). 

Using random sampling with a lot of masking makes it harder for the model to guess what's missing by just looking at the nearby patches. Finally, having very few patches allows us to design a more efficient system for processing the image, which we'll explain next. Finally, unmasked patches (25\% of the original image) will be used as input for a ViT encoder.

\subsection{Masked Decoder}

After the encoding phase, we have two sets of information: encoded data for the visible patches and special ``mask tokens''~\cite{bert}. Such mask tokens represent missing patches that the model needs to predict. Positional encoding is then added to both the encoded patches and mask tokens. 

This information is passed through another series of Transformer blocks, acting as a decoder. The decoder is only used during training to learn how to fill in the missing patches. It does this by predicting the actual pixel values for each masked area. The decoder's output is a vector of pixel values representing a patch, and the final step is to put all these patches back together to form a complete reconstructed image. 
To measure the goodness of the reconstruction process we compare the reconstructed image with the original one pixel by pixel. We use the mean squared error (MSE) to calculate this difference.

\subsection{Dataset Generation}

Our ultimate goal is to distinguish tumor tissue from healthy tissue and further classify tumor subtypes. This requires the model to learn informative representations of the tissues. To achieve this, we randomly extract a large number of unlabeled image regions from each WSI. This process aims to capture a diverse pool of tissue regions encompassing both tumor and non-tumor areas. \newline
The steps for extracting an image from WSI are listed below:

\textbf{Region Selection}. A square-shaped image patch is chosen from the whole slide image (WSI). The side length of the region is determined by sampling from a normal distribution. The mean and standard deviation of this distribution is calculated based on the size statistics of previously annotated ROIs in the BRACS dataset. \newline

\textbf{Region Quality Control}. The average variation is computed for the pixel intensities within the extracted patch. This is a measure of dispersion relative to the mean. If the average variation is greater than a predefined threshold, the region is considered informative and included in the dataset. This step aims to exclude regions from uninformative zones, such as borders, which often exhibit low variability with predominantly white or black pixels.

\subsection{Annotated ROIs classification}
This task investigates the ability of a pre-trained model to classify annotated regions of interest (ROIs) within whole slide images (WSIs) for tissue type classification. The approach is divided in two-steps: first, ROIs are fed into the pre-trained model, which utilizes MAE to generate informative feature vectors for the patches. Subsequently, a mean aggregation of these patch-level embeddings is performed in order to obtain a single vector for every ROI. These embeddings capture the essential characteristics learnt by the model from the data. Secondly, in order to carry out the classification task, we resort in a MLP with one hidden layer. The MLP utilizes the ROI embeddings as input features, where each class corresponds to a specific tissue type present within the ROIs.
This approach serves as an evaluation of the representativeness and discriminative power of the unsupervised embeddings generated by the pre-trained model. By successfully classifying different tissue types solely based on the embeddings, we can demonstrate that the learnt features effectively capture relevant biological information within the ROIs. 

\section{Experiments}
\begin{figure}[h]
  \centering
  \includegraphics[width=0.8\linewidth]{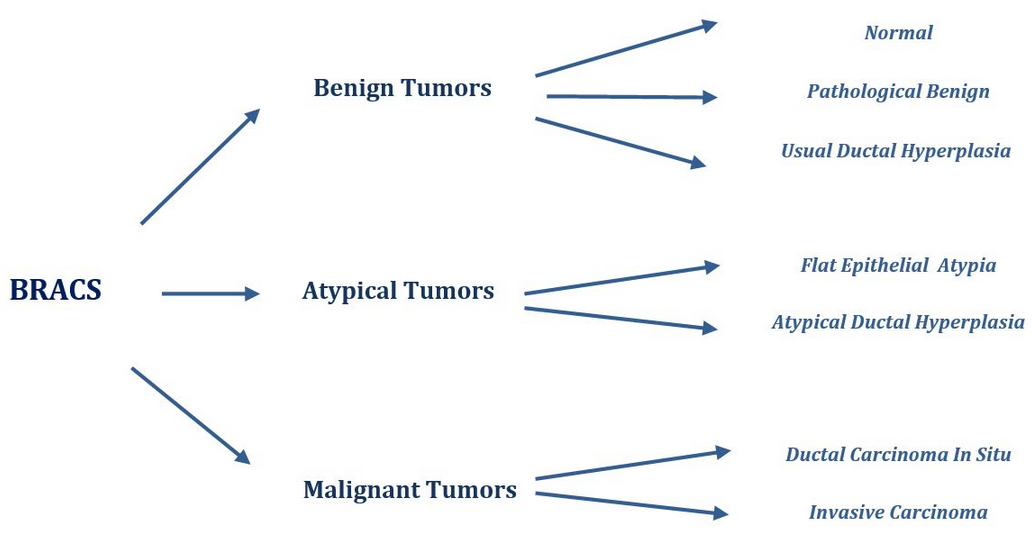}
  \label{fig:bracs_class}
  \caption{Classes in the bracs dataset}
\end{figure}

During the experimental phase we applied our method to breast cancer subtype classification using histopathological hematoxylin and eosin (H\&E) tissue images. Following the BRACS group's categorization (see Figure \ref{fig:bracs_class}), we designed two multi-class classification experiments. The first experiment differentiates cancerous from non-cancerous tissues, additionally incorporating an intermediate class for atypical cases. 

The second experiment extends the classification by attempting to categorize the tissue into seven distinct subtypes. Furthermore, we explore the utilization of a MAE network. Trained to reconstruct the image without prior labels, the model generates embeddings that will be analyzed. 

We will evaluate the clustering behavior of these embeddings in latent space and utilize attention maps to qualitatively assess the image regions on which the model focuses.
The results will be discussed in section \ref{con}.
\newline
\newline
\textbf{Generalization experiment.} We investigated the generalization capabilities of MAE embeddings for breast cancer histopathology image classification. The BACH dataset~\cite{bach}, containing 400 breast cancer histopathology images classified into four categories, was used for this evaluation. The pre-trained MAE model's encoder was used to extract feature representations (embeddings) from the BACH dataset images. Subsequently, a simple linear classifier was trained on the four classification labels specific to the BACH dataset.
\newline

\begin{figure*}[h]
  \centering
  \includegraphics[width=0.99\linewidth]{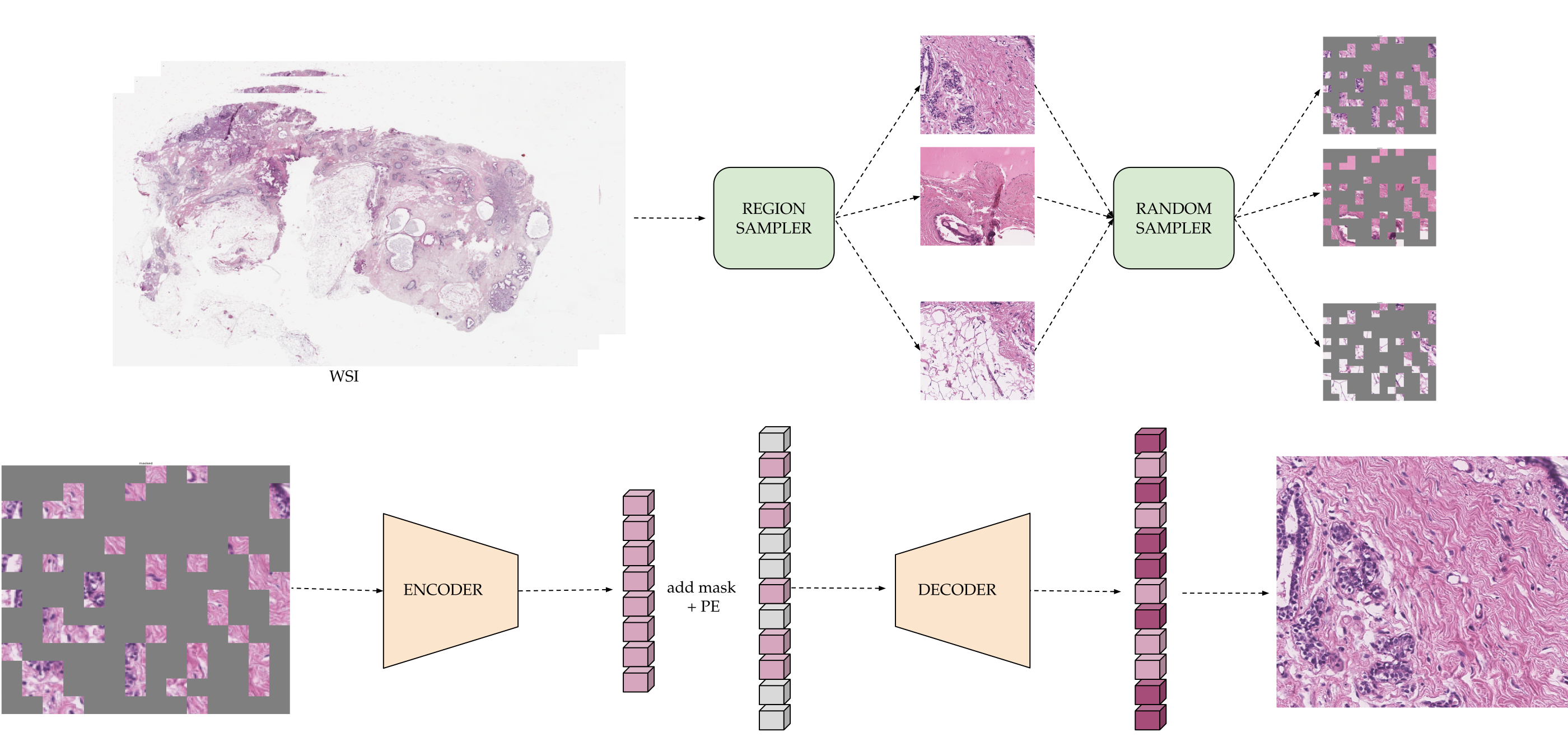}
  \label{fig:HMAE}
  \caption{\textbf{HMAE architetture}. (top) Tissue regions are randomly sampled from the original image. Subsequently, a random mask is applied, occluding 75\% of the image data. (down) The following autoencoder architecture receives the masked image as input and aims to reconstruct the hidden regions.}
\end{figure*}

\noindent
\textbf{Experimental setup.} The datasets were split into training, validation, and test sets using an 70/10/20 split. To account for potential variability, each classification experiment was run 100 times, and the average value for each chosen metric was reported. The MAE training was performed on a single Nvidia A40 GPU and took 32 hours. 


\subsection{Cancer classification}
This experiment evaluated our model's ability to classify tissue samples as cancerous or non-cancerous. Additionally, the model assigned ambiguous cases to an 'atypical tissue' class. For a standardized comparison, all models employed a ViT-S/16 encoder architecture. This facilitated a comprehensive evaluation against various baseline models, leveraging the groundwork established in \cite{acmil}. 
These models use either an attention mechanism or a ViT to correlate different region of a tissue image. They then classify the tissue by considering the relationships between these regions (the first two in the table perform the maximum and the average between the image patches respectively).
The model's performance, measured by F1-score and AUC, is presented in Table~\ref{tab:cls_res}.

\begin{table}[h]
\centering
\caption{Results of the other models compared to ours.}
\label{tab:cls_res}
\begin{tabular}{lll}
\hline
\multicolumn{1}{c}{\textbf{Model}} & \multicolumn{1}{c}{\textbf{F1-score}} & \multicolumn{1}{c}{\textbf{AUC}} \\ \hline
Max-pooling  & 0.596±0.029 & 0.823±0.033 \\
Mean-pooling & 0.522±0.038 & 0.739±0.007 \\
Clam-SB \cite{clam-sb}      & 0.631±0.034 & 0.863±0.005 \\
TransMIL \cite{transmil}     & 0.631±0.030 & 0.841±0.006 \\
DSMIL \cite{dsmil}        & 0.577±0.028 & 0.816±0.028 \\
DTFD-MIL \cite{dtfdmil}     & 0.612±0.080 & 0.870±0.022 \\
IBMIL \cite{ibmil}        & 0.645±0.041 & 0.871±0.014 \\
MHIM-MIL \cite{mhim-mil}     & 0.625±0.060 & 0.865±0.017 \\
ABMIL \cite{abmil}        & 0.680±0.051 & 0.866±0.029 \\
\textbf{ACMIL} \cite{acmil}                     & \textbf{0.722±0.030}                  & \textbf{0.888±0.010}             \\
\rowcolor[HTML]{FFCCC9} 
HMAE (ours) & 0.704±0.009 & 0.866±0.003 \\ \hline
\end{tabular}
\end{table}

\subsection{Sub-type Cancer classification}
This experiment seeks to develop a model capable of classifying diverse cancer subtypes within the regions of interest. This task presents a heightened challenge compared to prior endeavors due to the often subtle morphological distinctions between certain cancer types. Additionally, the inherent heterogeneity of the training data, encompassing a significant proportion of non-tumor tissue secondary to the random extraction process, introduces an inherent class imbalance within the latent representation space.
We have evaluated performance of various models based on the weighted F1 metric for a 7-class classification task (Table~\ref{tab:cls_res7}). 

\begin{table}[h]
\centering
\caption{Results of the other models compared to ours in sub-type cancer classification}
\label{tab:cls_res7}
\begin{tabular}{lll}
\hline
\multicolumn{1}{c}{\textbf{Model}} & \multicolumn{1}{c}{\textbf{F1-weighted}} \\ \hline
CLAM-MB/B~\cite{clam-sb}  & 0.548±0.010 \\
CGC-Net~\cite{gnn1}    & 0.436±0.005 \\
Patch-GNN~\cite{gnn3}    & 0.521±0.006 \\
TG-GNN~\cite{hactnet}      & 0.559±0.001 \\
CG-GNN~\cite{hactnet}     & 0.566±0.013 \\
HACT-Net~\cite{hactnet}        & 0.615±0.009 \\
TransPath~\cite{transpath}    & 0.567±0.02 \\
TransMIL~\cite{transmil}        & 0.575±0.007 \\
\textbf{ScoreNet}~\cite{scorenet}                  & \textbf{0.644±0.009} \\
\rowcolor[HTML]{FFCCC9} 
HMAE (ours) & 0.578±0.015 \\ \hline
\end{tabular}
\end{table}
Previous exposure to a larger proportion of non-cancerous tissue during the training phase appears to have influenced the model's prediction distribution. This is reflected in the superior F1-score achieved for the "normal" class compared to both the benchmark model and other classes within this investigation (see Table~\ref{tab:cls_res_nic}). However, performance on the remaining classes, particularly the "ic" class with the highest F1-score, remains comparable to previously tested models. 
These observations suggest that despite a potential bias towards non-cancerous tissue introduced by the training data, the model retains the ability to discriminate effectively between different cancer sub-types.
\begin{table}[h]
\centering
\caption{Single class classification F1-score (based on top-3 models in Table \ref{tab:cls_res7}).}
\label{tab:cls_res_nic}
\begin{tabular}{llll}
\hline
\multicolumn{1}{c}{\textbf{Label}} & \multicolumn{1}{c}{\textbf{ScoreNet}} & \multicolumn{1}{c}{\textbf{HACT-Net}} & \multicolumn{1}{c}{\textbf{Ours}} \\ \hline
Normal      & 0.646±0.022 & 0.616±0.021 & \textbf{0.683±0.022} \\ 
Benign      & \textbf{0.540±0.022} & 0.475±0.029 & 0.485±0.020 \\
UDH         & \textbf{0.484±0.022} & 0.436±0.019 & 0.445±0.070  \\
ADH     & \textbf{0.474±0.024} & 0.404±0.025  & 0.301±0.015 \\ 
FEA       & \textbf{0.779±0.007} & 0.742±0.014 & 0.702±0.018 \\
DCIS     & 0.629±0.020 & \textbf{0.664±0.026} & 0.633±0.019 \\
Invasive  & \textbf{0.910±0.014} & 0.884±0.002  & 0.893±0.015 \\
 \hline
\end{tabular}
\end{table}

\subsection{Generalization capabilities}
To evaluate the representational capacity of our model, we have investigated its ability to generalize to data coming from completly unseen WSIs. This has been achieved by assessing its performance on a classification task using a dataset (BACH) not included in the training phase. The BACH dataset contains ROIs from tumor and non-tumor tissues, further categorized into four distinct classes.
For consistent evaluation and to isolate the classification performance, the embeddings are fed into a linear classifier. Results are shown on Table~\ref{tab:cls_bach}.
\begin{table}[h]
\centering
\caption{Results using a linear classifier on the BACH dataset.}
\label{tab:cls_bach}
\begin{tabular}{lll}
\hline
\multicolumn{1}{c}{\textbf{Model}} & \multicolumn{1}{c}{\textbf{BRACS → BACH}} \\ \hline
HACT-Net            & 0.402±.028 \\
TransPath           & 0.618±0.048 \\
TransMIL            & 0.465±0.100 \\
CLAM-SB/B           & 0.575±0.036 \\
\textbf{ScoreNet}   & \textbf{0.734±0.035} \\
\rowcolor[HTML]{FFCCC9} 
HMAE (ours)        & 0.673±0.032 \\ \hline
\end{tabular}
\end{table}

\subsection{Qualitative Evaluation}
\begin{figure*}[h]
  \centering
  \includegraphics[width=0.7\linewidth]{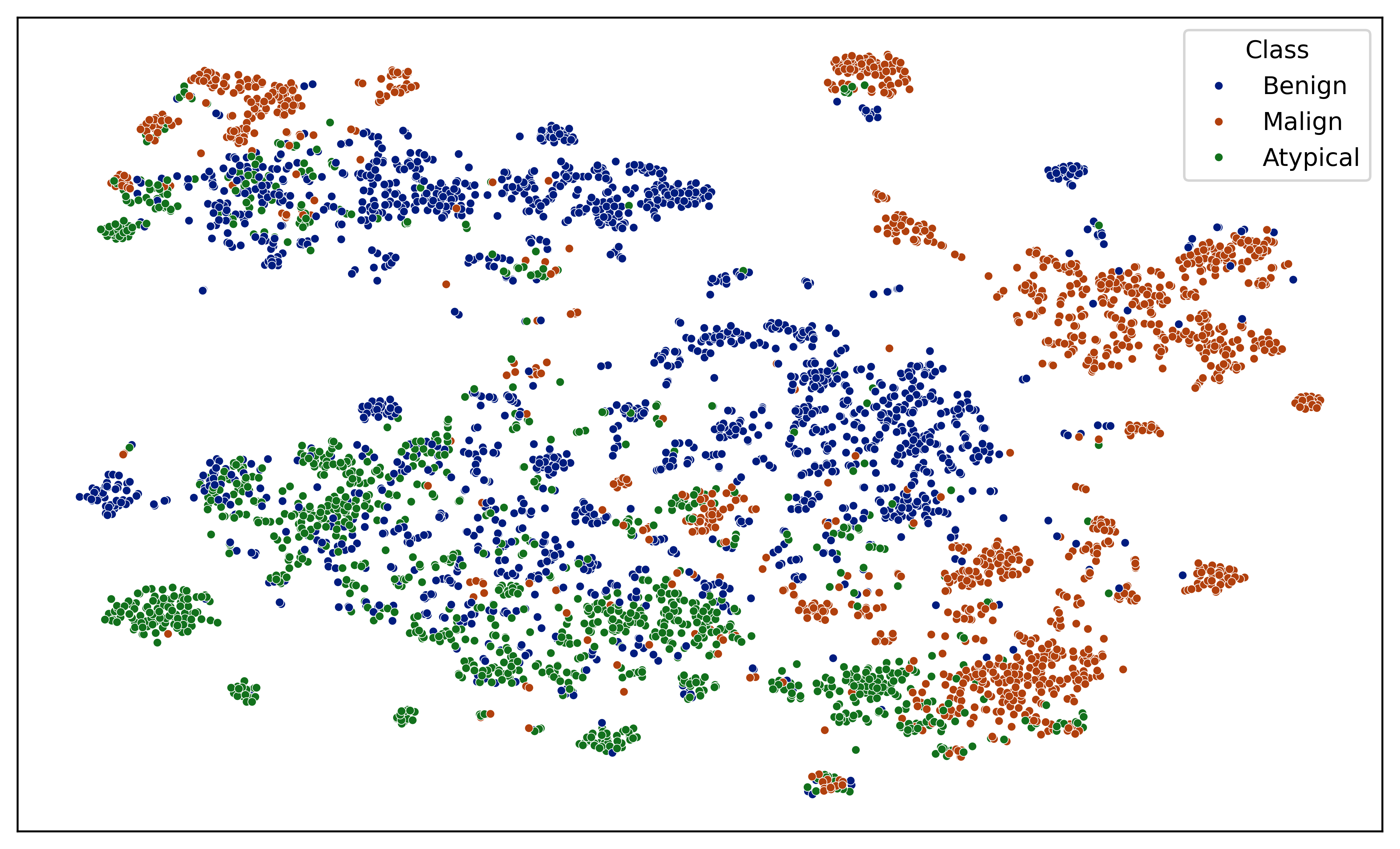}
  \caption{\label{fig:tsne}t-SNE of the RoI embeddings of the BRACS dataset. The sample are colored based on the classes.}
\end{figure*}

This section dives deeper into the quality of the learnt representations in the MAE model. We have employed t-SNE dimensionality reduction to visualize the embeddings in a 2D latent space, as shown in Figure~\ref{fig:tsne}. Each embedding is plotted and colored according to its corresponding class label. Interestingly, even though the model was trained unsupervised, some degree of class separation is already evident.

\begin{figure*}[h]
  \centering
  \includegraphics[width=0.99\linewidth]{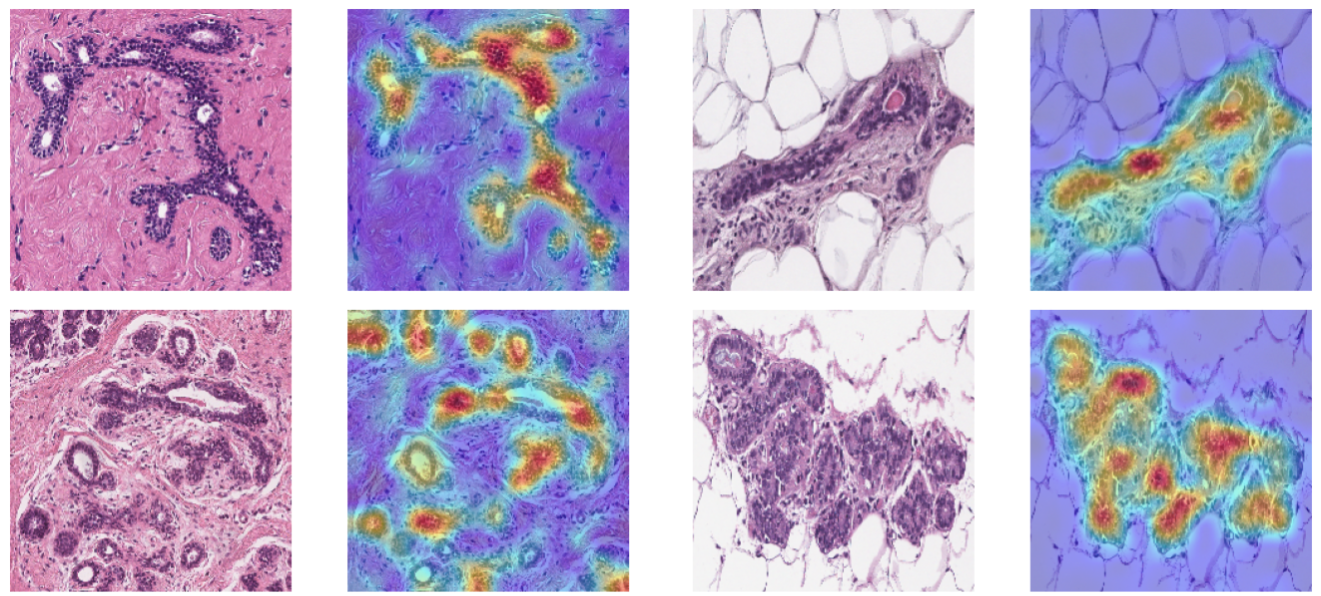}
  \caption{\label{fig:att_maps}Self attention heatmaps.}
\end{figure*}

Next, we have analyzed the attention maps, a crucial component of the Transformer architecture. These representations reveal which parts of the input image the model focuses on the most. By visualizing them, we can understand which image regions are most relevant for the model's predictions. Figure~\ref{fig:att_maps} showcases four original images alongside the attention maps generated by the final layer of the MAE. Notably, even during unsupervised training, the model appears to differentiate between connective and glandular tissue. This distinction likely arises because glandular tissue is structurally more complex, requiring the model to retain more information for reconstruction.

\section{Discussion and future works}\label{con}
In this work we presented an autoencoder architecture using a Vision Transformer (ViT) as the embedding module. This model effectively generates informative representations of input images by masking random regions and reconstructing the masked areas in a self-supervised learning setting. Applied to histopathological breast cancer images, the model successfully captures relevant features from both tumor and non-tumor regions. These learnt representations were then utilized as input to a classification model, achieving accurate cancer subtype identification.

Our model exhibits performance on par with current state-of-the-art techniques, consistently achieving second or third place in the conducted benchmarks. Notably, the latent space of the MAE is optimized solely for image reconstruction, and not specifically for classification tasks unlike the benchmark models. This observation strengthens the positive outcomes and underscores the efficacy of employing random input masking. Furthermore, the model's generalization capabilities suggest the potential for being applicable across diverse breast cancer datasets. 

Due to the high spatial resolution of whole slide images, data augmentation can be easily reachable. By extracting a larger number of random regions from each WSI, the dataset size can be significantly expanded while minimizing redundancy within the generated images. This step aims to achieve a more balanced representation of cancerous versus non-cancerous tissue within the training data without the need for large labeled datasets. 
\newline
In future work, we will explore the impact of expanding the training dataset to enhance the model's ability to differentiate cancerous and non-cancerous regions. Additionally, to assess the model's generalizability, we propose training it on a collection of diverse datasets. Finally, we also want to explore more in depth the dataset augmentation given by the random cropping of the WSIs by experimenting with more sophisticated methods that could potentially improve the model performance even more. 

    

\bibliography{sample-ceur}

\end{document}